\documentclass[lettersize,journal]{IEEEtran}
\usepackage{amsmath,amsfonts}
\usepackage{array}
\usepackage[caption=false,font=normalsize,labelfont=sf,textfont=sf]{subfig}
\usepackage{textcomp}
\usepackage{stfloats}
\usepackage{url}
\usepackage{verbatim}
\usepackage{graphicx}
\usepackage{cite}
\usepackage{hyperref}
\usepackage{array}  
\makeatletter
\newcommand{\thickhline}{%
    \noalign {\ifnum 0=`}\fi \hrule height 1pt
    \futurelet \reserved@a \@xhline
}
\usepackage{multirow}  
\usepackage{enumitem}  
\usepackage{makecell}  
\usepackage{tabularx}  

\usepackage{colortbl, booktabs}
\definecolor{light}{gray}{.85}
\definecolor{smooth}{gray}{0.95}
\definecolor{dg}{rgb}{0.0, 0.59, 0.09}

\usepackage{setspace}  

\usepackage{placeins}  

\usepackage{xcolor}  
\usepackage{amssymb} 

\usepackage{bm} 
\usepackage{booktabs}
\usepackage{diagbox}  

\usepackage{algpseudocode}
\usepackage[linesnumbered,ruled,vlined]{algorithm2e}

\usepackage{graphicx}  

\begin{document}

\title{Probabilistic Temporal Masked Attention for Cross-view Online Action Detection}

\author{Liping Xie,~\IEEEmembership{Member,~IEEE,} Yang Tan, Shicheng Jing, Huimin Lu,~\IEEEmembership{Senior Member,~IEEE,} Kanjian Zhang

\thanks{This work was supported in part by the National Natural Science Foundation of China under Grant 62372104, in part by the Guangdong Basic and Applied Basic Research Foundation under Grant 2022A1515110518, in part by the Research Fund of Key Program for Advanced Ocean Institute of Southeast University under Grant KP202402, in part by the Open Project of Anhui Provincial Key Laboratory of Multimodal Cognitive Computation, Anhui University, under Grant MMC202415. Besides, we thank the Big Data Computing Center of Southeast University for providing the facility support on the numerical calculations and the Taihu Lake Innovation Fund for the School of Future Technology of Southeast University. (Corresponding author: Liping Xie)}

\thanks{Liping Xie, Yang Tan, Shicheng Jing, Huimin Lu and Kanjian Zhang are with the Key Laboratory of Measurement and Control of Complex Systems of Engineering, Ministry of Education, School of Automation, Southeast University, Nanjing 210096, China (e-mail: lpxie@seu.edu.cn; yangtan@seu.edu.cn; scjing10@gmail.com; luhuimin@ericlab.org; kjzhang@seu.edu.cn).}

\thanks{Liping Xie, Yang Tan and Shicheng Jing are also with Anhui Provincial Key Laboratory of Multimodal Cognitive Computation, Anhui University, Hefei 230601, China (e-mail: lpxie@seu.edu.cn; yangtan@seu.edu.cn; scjing10@gmail.com).}

\thanks{Liping Xie, Huimin Lu and Kanjian Zhang are also with Advanced Ocean Institute of Southeast University, Nantong 226334, China (e-mail: lpxie@seu.edu.cn; luhuimin@ericlab.org; kjzhang@seu.edu.cn).}

\thanks{This paper has supplementary downloadable material available at http://ieeexplore.ieee.org., provided by the author. The material includes additional ablation studies and running explanations, providing further insights into our experimental setup and results. This material is provided as a PDF document and is 23.6MB in size.
}

}

\markboth{Probabilistic Temporal Masked Attention for Cross-view Online Action Detection}%
{Shell \MakeLowercase{\textit{et al.}}: A Sample Article Using IEEEtran.cls for IEEE Journals}

\maketitle

\begin{abstract}

    As a critical task in video sequence classification within computer vision, Online Action Detection (OAD) has garnered significant attention. The sensitivity of mainstream OAD models to varying video viewpoints often hampers their generalization when confronted with unseen sources. To address this limitation, we propose a novel \textbf{P}robabilistic \textbf{T}emporal \textbf{M}asked \textbf{A}ttention (PTMA) model, which leverages probabilistic modeling to derive latent compressed representations of video frames in a cross-view setting. The PTMA model incorporates a GRU-based temporal masked attention (TMA) cell, which leverages these representations to effectively query the input video sequence, thereby enhancing information interaction and facilitating autoregressive frame-level video analysis. Additionally, multi-view information can be integrated into the probabilistic modeling to facilitate the extraction of view-invariant features. Experiments conducted under three evaluation protocols—cross-subject (cs), cross-view (cv), and cross-subject-view (csv)—demonstrate that the PTMA achieves state-of-the-art performance on the DAHLIA, IKEA ASM, and Breakfast datasets.

    \end{abstract}

\section{Introduction}
\label{sec:introduction}

Online Action Detection represents a pivotal sequential online classification challenge, tasked with the frame-level classification of actions within streaming video data. This task is inherently complex, as it must be performed without the benefit of future information, making it highly relevant for surveillance systems \cite{Pavlidis_2001_Urban}, human-computer interaction \cite{Liu_2022_EHPE, Xie_2021_Graph}, robot imitation \cite{Fang_2023_HODN, Liu_2018_Imitation}, and autonomous driving \cite{Yu_2020_Spatio, Liu_2023_Petrv2}.

\begin{figure}[t]
    \vspace{-0.2cm} 
    \centering
    \includegraphics[width=0.48\textwidth]{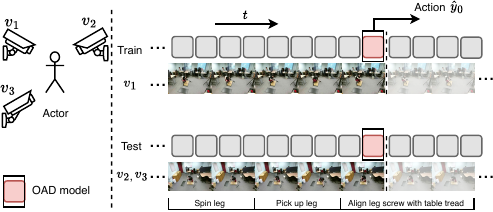}
    \caption{Illustration of OAD under cross-view settings. The OAD model is trained on $v_1$ while tested on unseen $v_2$ and $v_3$. The training and testing need not contain identical footage.}
    \label{fig:cvoad}
    \vspace{-0.3cm} 
  \end{figure}

Unlike traditional tasks like action localization \cite{Zhang_2024_Integration, Xia_2023_Exploring, Ju_2022_Adaptive}, recognition \cite{Zhu_2024_Part, Wang_2024_CLIP}, and detection \cite{Gan_2022_Temporal} that benefit from global video perception, OAD faces greater challenges due to its local perception and lack of future information access. Current OAD approaches predominantly use mixed camera views for training, largely overlooking view-specific cues. While Tan \emph{et al}. \cite{Tan_2024_Annealing} have explored multi-view training, their focus on joint-view regularities neglects the crucial challenge of generalizing to unseen views \cite{Xie_2020_Efficient, Leng_2023_Online, Guo_2022_Uncertainty, Eun_2021_Temporal, Xie_2018_Early}. As shown in Figure \ref{fig:cvoad}, models trained on view $v_1$ often fail in unseen views $v_2$ and $v_3$, highlighting the limitations of current training paradigms. This gap is particularly critical in real-world applications like bank surveillance \cite{Zhang_2019_Multi}, where cross-view generalization is essential for operational effectiveness, emphasizing the need for more robust cross-view OAD models.

Drawing from common intuition, it is challenging for a pre-trained model to generalize scenarios it has not encountered before. However, in cross-view settings, the unseen viewpoint is not entirely foreign but rather a variant of a similar scenario. This implies that the cues learned from the seen viewpoints can potentially aid in generalizing to the unseen viewpoints.  Therefore, recognizing the inherent similarities in action context across different video viewpoints, we hypothesize that a shared probability distribution underlies the generation of video features, despite minor variations \cite{Zhang_2024_novel, Shah_2023_Multi, Das_2023_ViewCLR}. It is suggested that a low-dimensional latent representation derived from a single viewpoint can capture essential view-level information, which can be employed to represent partial unseen view-level information. This approach, when applied, can enhance the model's ability to detect actions consistently across varied viewpoints, potentially optimizing performance in cross-view OAD scenarios \cite{Xue_2023_Learning, Gao_2021_View}.

Bayesian-informed probabilistic models are adept at extracting the latent distributions from observational data, offering significant utility for analytical tasks \cite{Xu_2022_Probabilistic, Guo_2022_Uncertainty}. Variational Autoencoder (VAE), characterized by the encoder-decoder structure, efficiently maps complex observational data into a low-dimensional probabilistic latent space, allowing for both navigation and reconstruction of the original data \cite{Xu_2023_Conditional, Tong_2023_Probabilistic}. Building upon these principles, we propose the Probabilistic Temporal Masked Attention (PTMA) model, which consists of two branches: a probabilistic branch that employs a VAE to derive latent frame-level encodings and a classification branch that utilizes these encodings in a GRU-TMA cell to perform temporal masked attention and autoregressive classification. The probabilistic branch enriches the video features with refined view-level information, facilitating OAD across different views. However, this process solely incorporates a single viewpoint and does not ensure that the learned latent encoding can fully represent other viewpoints. Consequently, when the data source encompasses multiple viewpoints, such as $v_1$ and $v_2$, we leverage $v_1$ as the input for the probabilistic branch and $v_2$ as the reconstruction label for the decoder (\emph{i.e.}, $v_{12}$-$v_3$ setting), thereby self-supervising the training of the probability model. The probabilistic encoder is capable of capturing view-invariant features that span across two viewpoints under such circumstances, resulting in a richer representation than what is achieved with the latent distribution. In the classification branch, a temporal mask is generated to prevent the distant history from interfering with OAD during inference. The PTMA model utilizes a Multi-Layer Perceptron (MLP) layer that maps the temporal encodings from the GRU-TMA cell to the action classification space. Besides, different evaluation mechanisms are adopted to assess the generalization capability of OAD models. Exhaustive experiments and ablation studies demonstrate the robustness and effectiveness of PTMA in cross-view OAD scenarios.

\textbf{Our contributions can be summarized as follows:}
\begin{itemize}[label=\tiny$\bullet$]
    \item Exploring the complexities inherent in OAD tasks within cross-view scenarios for the first time, and assess performance through three distinct evaluations.

    \item Introducing the Probabilistic Temporal Masked Attention (PTMA) model, which features a probabilistic branch to extract latent view-level representations and a parallel classification branch for accurate action classification. Multi-view sources are leveraged in the probabilistic branch to learn view-invariant features, enhancing the model's generalization.
    
    \item Integrating a GRU-TMA cell within PTMA's classification branch, in which the TMA module adeptly refines encodings by mitigating the impact of distant history and enhancing the interaction between latent representations and raw temporal encodings in GRU.
    
    \item Demonstrating PTMA's superiority against the prevailing mainstream OAD models through extensive experiments on the DAHLIA, IKEA ASM, and Breakfast datasets, essentially achieving state-of-the-art results.
    \end{itemize}

\section{Related Work}

\subsection{Online Action Detection}
Online action detection was first introduced by De Geest, \emph{et al}. \cite{DeGeest_2016_Online}, aiming at classifying untrimmed video streams that are input online. Current methods involve two main types of architectures: RNNs \cite{Xu_2019_Temporal, Eun_2020_Learning, Kim_2021_Temporally, An_2023_MiniROAD} and Transformers \cite{Wang_2021_OadTR, Xu_2021_Long, Chen_2022_GateHUB, Zhao_2022_Real, Wang_2023_Memory}. Within the RNN architecture, TRN \cite{Xu_2019_Temporal} modifies the internal structure of RNNs to model context using historical evidence and predicted future information. IDU \cite{Eun_2020_Learning} retains information as history based on its relevance to the current action, facilitating feature representation. MiniROAD \cite{An_2023_MiniROAD} employs non-uniform weights to calculate loss, allowing RNN models to learn from predictions during training as if in an inference phase. 
In the Transformer architecture, OadTR \cite{Wang_2021_OadTR} was the first to apply the Transformer to the OAD task. LSTR \cite{Xu_2021_Long} models use long-term and short-term memory mechanisms to model extended sequence data. MAT \cite{Wang_2023_Memory} employs a novel memory-based prediction paradigm to simulate the entire temporal structure, including past, present, and future. HCM \cite{Liu_2024_HCM} selects challenging clips for optimization and employs loss functions to enhance intra-class compactness and inter-class separation. Models based on the Transformer architecture are typically challenging to train and often have large GFLOPs. Therefore, we construct the PTMA model based on the RNN architecture. Furthermore, datasets typically utilized in conventional OAD tasks (\emph{e.g.}, THUMOS'14) often contain videos with a single or just a few actions, which contrasts with the complexity found in real-world scenarios. In this paper, we concentrate on the cross-view OAD task employing more complex and realistic datasets (\emph{e.g.}, IKEA ASM) to address these challenges.

\subsection{Cross-view Action Understanding}
Significant works have been done in the realm of action understanding across varying viewpoints \cite{Siddiqui_2024_DVANet, Marsella_2021_Adversarial}. DA-Net \cite{Wang_2018_Dividing} integrated prediction probabilities from different viewpoints and used them as weighted scores for action recognition. The actions featured in OAD scenarios are more complex, and weighted scores cannot simply represent the importance of view-level features. Models such as ATSCL \cite{Tan_2024_Annealing} and ViewCLR \cite{Das_2023_ViewCLR} chose to use contrastive learning strategies for feature fusion between different views. Haresh \emph{et al.} \cite{Haresh_2021_Learning} designed a Soft-DTW loss based on multi-view videos to obtain better feature extraction. However, the cross-view OAD task requires that different viewpoints be input simultaneously at the same time during testing, resulting in the ineffectiveness of the above methods. Piergiovanni \emph{et al.} \cite{Piergiovanni_2021_Recognizing} directly rendered poses from different viewpoints in the world coordinates to assist in action recognition, but this is equivalent to introducing other information rather than directly utilizing the original video information. Ghoddoosian \emph{et al.} \cite{Ghoddoosian_2022_Weakly} only considered multi-view information assistance for OAD, rather than considering OAD applications in cross-view scenarios. Despite these notable contributions, there are considerable differences between these tasks and OAD, prompting the development of our PTMA model. So far, cross-view OAD is still a problem to be studied. Thus, PTMA is designed to integrate video features in a cross-view manner, offering a robust solution for cross-view OAD tasks and distinguishing itself within the current landscape of mainstream models.

\subsection{Variational Autoencoder}
In the field of video understanding, there has been extensive research on probabilistic modeling \cite{Zhou_2019_Recognizing, Srivastava_2021_variational}. ACT-VAE \cite{Xu_2023_Conditional} maintained temporal coherence based on action categories and establishes a novel temporal model in the latent space. Guo \emph{et al}. \cite{Guo_2022_Action2video} developed a temporal variational autoencoder to enhance the diversity of output action poses, assisting in 3D action generation. Panousis \emph{et al}. \cite{Panousis_2021_Variational} utilized Variational Bayes for tractable inference while inferring time-varying dependencies through latent variables for action recognition. Wei \emph{et al}. \cite{Wei_2023_Unsupervised} developed TranSVAE to model the encoding of static and dynamic video information for unsupervised video domain adaptation. In this paper, we model the latent action representation with a variational autoencoder and apply it to learn cross-view information from unseen viewpoints.

\section{Methodology}
\label{sec:methodology}
This section begins with a definition of the cross-view OAD task. We then succinctly introduce the proposed PTMA model, emphasizing its core components and their roles in addressing the task's challenges. We proceed to explain the probabilistic modeling foundation of PTMA and describe the GRU-TMA cell's function, particularly its use of a temporal mask to reduce the impact of distant history on inference. Multi-view sources are introduced to enhance the reconstruction of video inputs for better view-level representations if the number of viewpoints allow for this approach. The section concludes with a brief overview of our training paradigm used to prepare PTMA for cross-view OAD.

\subsection{Task Formulation}
OAD aims to detect actions in untrimmed video streams under real-time conditions. In a cross-view setting, the input video stream exhibits a distinct viewpoint bias, as the model cannot capture information beyond a specific view. Therefore, cross-view OAD can be defined as training with a given view $ v_i $ and its corresponding video features $ \mathbf{X}^{v_i} = \{x_t^{v_i} \}_{t=-K_i+1}^{t=0}$, and testing on $ \mathbf{X}^{v_j} = \{x_t^{v_i} \}_{t=-K_j+1}^{t=0} $ from an unseen view $ v_j $, in which $K_i$ and $K_j$ are the numbers of video frames. Here, the frame-level feature $x_{t}^{v_i}$ and $x_{t}^{v_j}$ are both with dimension of $D$. In the training phase, frames are sliced into multiple segments using a sliding window of length $T$ and input into the model to learn the RGB and motion patterns in the video. Additionally, we denote the classification of the current frame $ f_0 $ as $ y_0 $, where $ y_0 \in \{0, 1, ..., C\} $, with 0 indicating the background and $ C $ representing the total number of action categories. Thus, the model's prediction is framed as the probability distribution $ p(y_{0}|\mathbf{X}^{v_j}, \theta^{*}) $. Here, $ \theta^{*} = \mathrm{argmax\ } p(y_0|\mathbf{X}^{v_i}, \theta) $. In the following narrative, $v_i$ and $v_j$ are omitted if differentiation between them is not required.

\subsection{Model Overview}

\begin{figure*}[ht]
    \vspace{-0.2cm} 
    \centering
    \includegraphics[width=1.\textwidth]{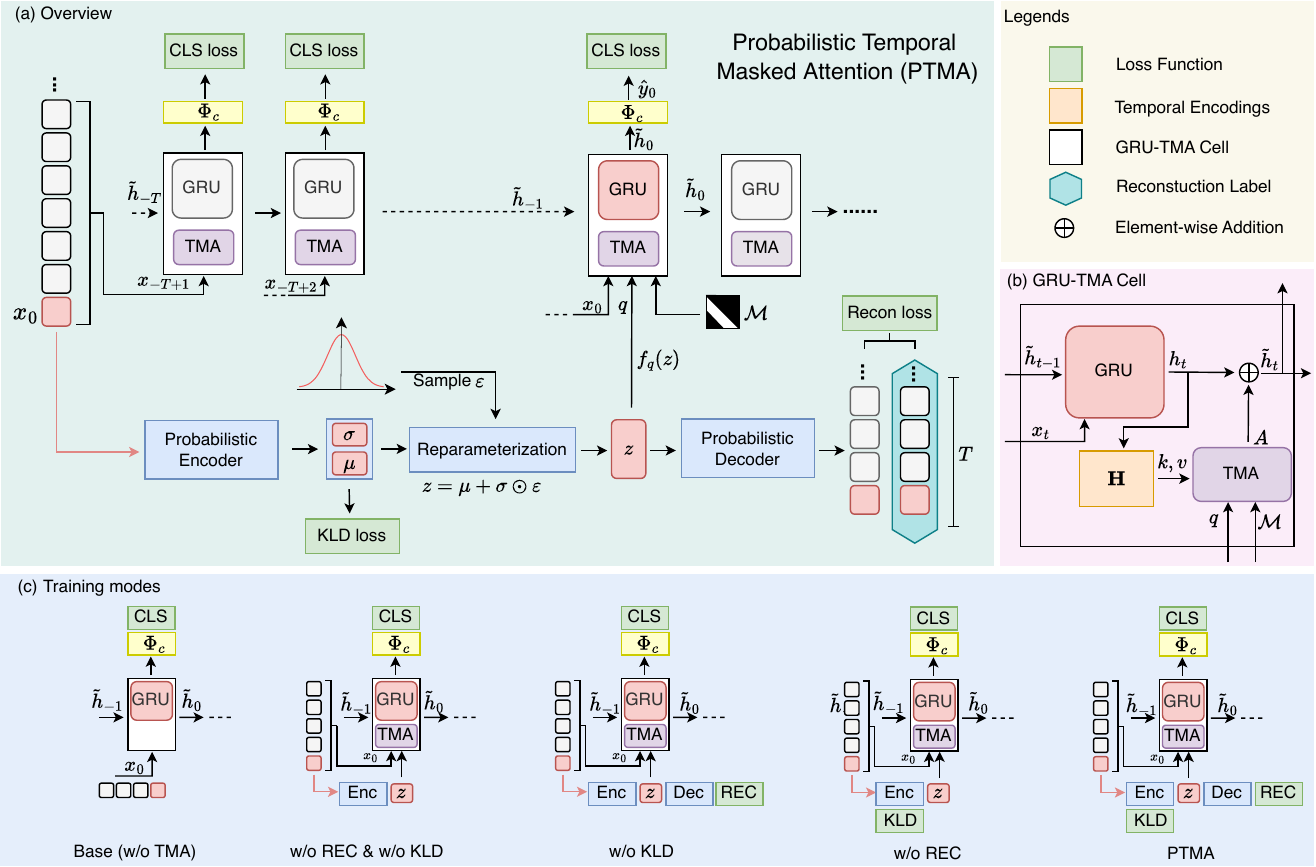}
    \caption{Illustration of the proposed PTMA model. (a) PTMA's overview. The upper classification branch includes a GRU-TMA cell for preliminary encoding input features $\mathbf{X}$. The lower probabilistic branch acts as a probabilistic model, mapping $\mathbf{X}$ to latent representations $\mathbf{z}$, which is used to interact with the GRU-TMA cell for temporal masked attention. (b) The expanded view of the GRU-TMA cell. The preliminary encoding $\mathbf{H}$ and the attention output $\mathbf{A}$ are element-wise added to produce the refined temporal encoding $\tilde{\mathbf{H}}$ for classification. (c) Different training modes for PTMA. Various modules of PTMA are activated or deactivated and the base model is a fine-tuned GRU without TMA module.}
    \label{fig:overview}
  \end{figure*}

In Figure \ref{fig:overview}, we illustrate the architecture of the proposed PTMA model, which is bifurcated into two primary branches: the classification branch above and the probabilistic branch below. The classification branch capitalizes on the GRU to facilitate rapid recursive inference for OAD. However, to address the limitations of relying on the classification branch alone, which might neglect the nuances of view-specific features, the probabilistic branch integrating a Probabilistic Model is considered. This model compresses view-level features into a latent view-specific encoding, denoted as $\mathbf{z}$. This encoding serves as a query within the Temporal Masked Attention (TMA) mechanism of the classification branch. It enables the extraction of essential information from novel viewpoints, thereby refining the model's raw temporal encodings in cross-view OAD tasks. To ensure that this process is dynamically integrated at each time step, we merge the TMA module with the GRU, creating what we term the GRU-TMA cell. This fusion allows for the continuous and adaptive refinement of OAD in response to incoming video data across different views. Suppose that the cell is $\mathbf{\Phi}_d$ and the classifier behind it is $\mathbf{\Phi}_{c}$, from which we get $\tilde{\mathbf{H}}=\mathbf{\Phi}_{d}(\mathbf{X}_{-T+1:0})$ and $\hat{y}_0=\mathbf{\Phi_{c}}(\tilde{\mathbf{H}})$. 

\subsection{Probabilistic Modeling}
\textbf{Probabilistic Assumption.} In this paper, the assumption is made that the video features of the same action captured from different views are identically distributed, expressed as 

\begin{equation}
\label{eq:assumption_1}
\mathbf{X}^{v_i}, \ \mathbf{X}^{v_j} \sim p(\mathbf{z}),
\end{equation}
where the latent variable $\mathbf{z}$, assumed to follow a Gaussian distribution, represents the underlying latent encoding shared in different views. If a latent encoding $\mathbf{z}$ can effectively reconstruct viewpoint $v_i$, then we believe it can also represent partial features of viewpoint $v_j$.

\textbf{Probabilistic Inference.} A high-quality latent encoding $\mathbf{z}$ should maximize the likelihood of the input data $p(\mathbf{X})$. We use the maximization of $ \log p(\mathbf{X}) $ as a substitute for the maximization of $ p(\mathbf{X}) $, and it is equivalent to Eq. \ref{eq:log_px}:

\begin{equation}
\label{eq:log_px}
\log p(\mathbf{X}) = \int_{\mathbf{z}} p(\mathbf{z}|\mathbf{X}) \log p(\mathbf{X}) d\mathbf{z}.
\end{equation}

The high dimensionality of the latent variable $\mathbf{z}$, which encapsulates the complexity of video features, renders the direct computation of the conditional probability $p(\mathbf{z}|\mathbf{X})$ infeasible. To address this, we adopt a variational approach, employing a Gaussian distribution $q(z|x)$ to approximate the true posterior $p(z|x)$. This approximation allows us to estimate the data likelihood $p(\mathbf{X})$ as accurately as possible. To optimize the latent representation $\mathbf{z}$ that best explains the observed data, we parameterize the variational distribution $q_{\phi}(z|x)$ as a Gaussian $\mathcal{N}({\mu}, \sigma^2)$. Due to the temporal motion relationship between adjacent video frames, their generation process is not independent. Therefore, we use the average distribution of video frames within a temporal window to replace the original probability distribution of the video frames. This parameterization shifts the focus to refining the parameters ${\mu}$ and ${\sigma}^2$, which are depicted in Eq. \ref{eq:approximate} as the optimization objective:

\begin{equation}
\begin{aligned}
    \log p(\mathbf{X}) & \approx \frac{1}{T} \sum\limits_t \log p(x_{t}) \\
    &=\frac{1}{T} \sum\limits_t \int_{z} q_{\phi}(z|x_{t}) \log p(x_{t}) dz.
    \label{eq:approximate}
\end{aligned}
\end{equation}

In alignment with \cite{Kingma_2013_Auto, Doersch_2016_Tutorial}, the final result of $\log p(\mathbf{X})$ is equivalent to Eq. \ref{eq:elbo_kl}:

\begin{equation}
    \label{eq:elbo_kl}
    \begin{aligned}
    \log p(\mathbf{X}) &\approx
    \frac{1}{T} \sum\limits_{t=1}^{T}  \left[
    \int_{\mathbf{z}} q_{\phi}(\mathbf{z}|\mathbf{X}_{t})
     \log \left(\frac{p_{\theta}(\mathbf{X}_{t}|\mathbf{z})p(\mathbf{z})}{q_{\phi}(\mathbf{z}|\mathbf{X}_{t})}\right) d\mathbf{z} \right. \\
    &\quad \left. + KL(q_{\phi}(\mathbf{z}|\mathbf{X}_{t}) \| p(\mathbf{z}|\mathbf{X}_{t})) \right],
    \end{aligned}
\end{equation}
where $p_{\theta}$ represents the parameterized generative process from the latent space.

The second term on the right-hand side of Eq. \ref{eq:elbo_kl} denotes the Kullback-Leibler Divergence (KLD) between the true posterior $p(z|x_{t})$ and the variational approximation $q_{\phi}(z|x_{t})$, which is inherently positive. This implies that the first term constitutes the Evidence Lower Bound (ELBO) for $\log p(\mathbf{X})$. Upon simplification, the expression presented in Eq. \ref{eq:ELBO} can be derived:

\begin{equation}
\begin{aligned}
\log p(\mathbf{X}) &\geqslant \frac{1}{T} \sum_{t} \left\{
    -D_{KL}\left[q_{\phi}(z|x_{t}) \| p(z)\right]\right. \\
    &\quad \left. + \mathbb{E}_{q_{\phi}(z|x_{t})}\left[\log p_{\theta}(x_{t}|z)\right] \right\},
\end{aligned}
\label{eq:ELBO}
\end{equation}
where $D_{KL}$ signifies the KL divergence. The items on the right-hand side are KLD loss $\mathcal{L}_{kld}$ and reconstruction loss $\mathcal{L}_{rec}$, respectively.

The pursuit is to engineer the latent variable $\mathbf{z}$'s distribution to achieve a dual minimization: the reconstruction loss of the input features $\mathbf{X}$ and the KLD loss against the true posterior, ensuring the model is tightly calibrated to the data's underlying structure. Within the sophisticated framework of probabilistic inference, the model transforms input $\mathbf{X}$ into the defining parameters of a Gaussian distribution—specifically, the mean $\bm{\mu}$ and standard deviation $\bm{\sigma}$. These parameters are pivotal for distilling the latent action encodings, a process we have masterfully integrated using MLPs. This approach not only streamlines the encoding process but also elevates the model's ability to capture the nuanced dynamics of cross-view OAD.

\begin{figure}[ht]
    \centering
    \includegraphics[width=0.45\textwidth]{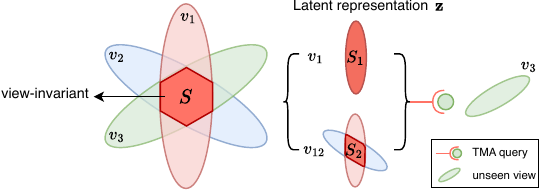}
    \caption{Comparison of latent space for probabilistic modeling using single-view or multi-view sources for data reconstruction. The area $S$ is an ideal latent space where the model can learn view-invariant features.}
    \label{fig:latent_space}
    \vspace{-0.2 cm} 
  \end{figure}

\textbf{Probabilistic Generation.}
The probabilistic generation within PTMA encompasses latent action encodings from $q_{\phi}(z|x_t)$ during inference. These encodings are subsequently fed into a probabilistic decoder designed to accurately reconstruct the original video features. By minimizing the reconstruction loss, it can be ensured that the  $\mu_t$ and $\sigma_t$ generated at time $t$ can lead to a latent view-specific representation of the video features. This representation encapsulates the essence of unseen videos across diverse views, allowing the model to understand actions from a known viewpoint when confronted with an unseen viewpoint. However, this process involves only a single viewpoint, which is insufficient for latent representations to capture the shared features of unseen viewpoints as Figure \ref{fig:latent_space} shows. To address this issue, multi-view sources can be introduced to supervise the training of probabilistic decoder. In the original mechanism, the output reconstruction of the probabilistic model $\mathbf{X}^{(r)}$ is anticipated to align with the input $\mathbf{X}^{v_i}$. However, in this scenario, the decoder is expected to reconstruct features from another viewpoint $\mathbf{X}^{v_k}$. In this case, the latent representation is trained to be more view-invariant (\emph{i.e.}, area $S_2$ vs. $S_1$), which is crucial for the generalization of unseen viewpoint $v_j$.

We implement the reparameterization technique to facilitate a stable and differentiable sampling process. This involves a random variable vector $\bm{\varepsilon}$ from $\mathcal{N}(\mathbf{0},\mathbf{I})$, which is then rescaled and shifted to align with the desired Gaussian distribution parameters—mean $\bm{\mu}$ and standard deviation $\bm{\sigma}$ as depicted below:

\begin{equation}
\begin{aligned} 
    \mathbf{z} = & \bm{\mu} + \bm{\sigma} \odot \bm{\varepsilon}, \\
     \bm{\varepsilon} & \sim \mathcal{N}(\mathbf{0},\mathbf{I}),
\end{aligned}
\end{equation}
where $\mathbf{z},\ \bm{\mu},\ \bm{\sigma}$ and $\bm{\varepsilon} \in \mathbb{R}^{T \times D_z}$, $D_z$ is the dimension of the latent variable.

Following the sampling of the latent variable $\mathbf{z}$, a probabilistic decoder comprised of an MLP is leveraged to reconstruct the input pattern. The output $\mathbf{X}^{(r)}$ matches the dimensionality of the input, directly contributing to the mathematical expectation term in the ELBO formulation (Eq. \ref{eq:ELBO}). This synthesis of probabilistic modeling and decoder reconstruction is central to the effectiveness of the PTMA model in cross-view OAD.

\subsection{Temporal Masked Attention}
For the independent classification branch, training with a fixed view tends to focus the model's attention on specific view-level features, significantly reducing its generalization across other views. Therefore, incorporating other view-level features (\emph{i.e.}, the latent representation $\mathbf{z}$) is crucial for OAD on unseen views. To address this, the GRU is combined with temporal masked attention as a GRU-TMA cell in the classification branch, intended to use the latent representation learned from the seen viewpoint to query the unseen viewpoint.

The raw temporal encodings $\mathbf{H}_{-T+1:0}$ of GRU are inherently tailored to capture frame-specific occurrences and are often view-dependent. To transcend this limitation, we introduce latent variable $\mathbf{z}$ as queries and utilize raw encodings as keys and values for querying, as depicted in Eq. \ref{eq:tma}. The output of attention serves as auxiliary information for joint OAD on the test view.
\begin{equation}
\label{eq:tma}
\begin{aligned}
    \mathbf{A} &= \operatorname{\mathcal{S}oftmax}\left[\frac{{\mathbf{F}}_z  \left({{\mathbf{H}}}\right)^{\top}}{\sqrt{\alpha}} + {\mathcal{M}} \right] \cdot {\mathbf{H}},
\end{aligned}
\end{equation}
in which ${\mathbf{F}}_z=f_q(\mathbf{z})$ represents the expansion of latent variables $\mathbf{z}$ into the encoding space of the GRU's output and $f_q$ denotes the mapping $\mathbb{R}^{T \times D_z} \rightarrow \mathbb{R}^{T \times D}$. $\alpha$ serves as a hyperparameter to adjust the intensity of attention. $\mathcal{M}$ is a temporal mask designed to mitigate the influence of distant historical information during testing, and more details will be discussed in Section \ref{sec:temporal_mask}.

\begin{figure*}[ht]
    \vspace{-0.2cm} 
    \centering
    \includegraphics[width=0.9\textwidth]{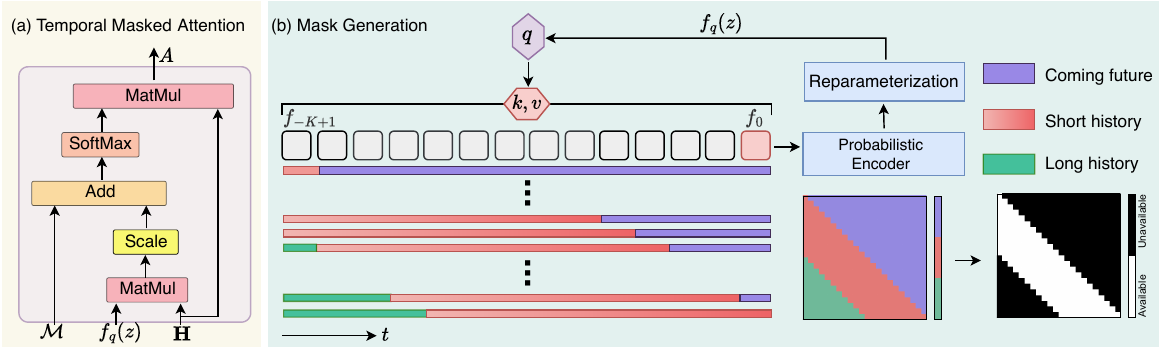}
    \caption{Illustration of the Temporal Masked Attention. (a) Computation of TMA. The input $f_q(z)$ functions as query ($q$), and the raw temporal encoding $\mathbf{H}$ is key ($k$) and value ($v$). (b) Generation of the temporal mask. The red square denotes the temporal window for OAD, spanning a length of T. The purple area symbolizes an inaccessible future, and the green area indicates a distant long history. During inference, where the temporal window extends over the full video sequence, the inclusion of distant history could hamper the current OAD task. To counteract this, information exchange with latent view-level representation is confined to the red zone, aligning with training constraints. The model is guided to prioritize short-term historical information by masking both future and distant history frames, thereby enhancing task relevance and model focus.}
    \label{fig:tma}
  \end{figure*}

The refined feature encodings, achieved via the element-wise addition of the raw feature encodings and the temporal masked attention output, serve as the input to the classification layer for final action prediction, as shown in Eq. \ref{eq:prediction}:

\begin{equation}
    \label{eq:prediction}
    \begin{aligned}
        \hat{y} = \mathbf{\Phi}_c(\tilde{\mathbf{H}})= \mathbf{\Phi}_c ({\mathbf{H}} \oplus  {\mathbf{A}}).
    \end{aligned}
\end{equation}

\subsection{Temporal Mask}
\label{sec:temporal_mask}
In the model's training phase, a fixed time window of length $T$ is utilized for training, whereas during inference, the model operates in an autoregressive manner to process the entire video sequence from end to end. In this condition, the interaction between the model's branches can sometimes lead to less-than-ideal effects, where attention from the distant past may negatively impact the current OAD task.

In a typical setup, to avoid future information leakage, an attention mask matrix is used with the upper triangular area set to $-\infty$, which masks out future timesteps. Yet, when the video length $K$ exceeds the window length $T$, it becomes impractical to factor in frames that are more than $T$ time steps away. Therefore, the attention mask should be formulated to reflect this constraint:
\begin{equation}
    \mathcal{M}_{ij}=\left\{\begin{array}{ll}-\infty,&-K+1<j\leqslant i-T\:or\:j>i, \\\\0,&i-T<j\leqslant i,\end{array}\right.
    \label{eq:temporal_mask}
\end{equation}
in which $\mathcal{M}_{ij}$ is the mask of $q_i$ querying $h_{j}$. In response to this requirement, a temporal mask is introduced as shown in Figure \ref{fig:tma}.

The temporal mask involves strategically limiting the impact of historical information on the current state by overlaying and modulating the attention information with a designed mask. This method effectively protects the model from interference by frames from the distant past, which can be particularly beneficial for maintaining high accuracy in OAD tasks. By performing a pixel-wise addition between the mask matrix and the attention matrix, the model integrates the mask in an efficient manner that allows for parallel computation, streamlining the overall OAD process.


\subsection{Training Paradigm}

In this paper, we input features $\mathbf{X}$, sampled with a window length of $T$, into the GRU to obtain raw feature encodings $\mathbf{H}$. Simultaneously, the query result of the latent variable $\mathbf{z}$ is applied to produce refined encodings $\tilde{\mathbf{H}}$, the output of the GRU-TMA Cell. A linear classification layer is used to classify all frames within the temporal window. The pseudocode of the detailed training paradigm can be found in Algorithm \hyperlink{alg:ptma_training_alg}{1}.

The PTMA's cross-view classification is supervised by a cross-entropy loss, depicted as

\begin{equation}
\label{eq:cls}
\mathcal{L}_{cls}=-\sum_{t=-T+1}^{0} \sum_{i=0}^{C} y_{t_{i}} \log \hat{y}_{t_{i}}.
\end{equation}

Here, $y_t$ is the groundtruth, and $\hat{y}_t$ is the predicted label at frame $f_t$.

To achieve higher-quality and more robust compressed representations $\mathbf{z}$, we derive the reconstruction loss $ \mathcal{L}_{\text{rec}} $ for rebuilding video features and the KLD loss $\mathcal{L}_{\text{kld}}$ to approximate the true distribution $p(\mathbf{z})$ based on Eq. \ref{eq:ELBO} as follows:

\begin{equation}
\begin{aligned}
    \label{eq:kld_rec}
    \mathcal{L}_{rec}
    &=\frac{1}{T} \sum_{t=-T+1}^{0} \sum_{i=1}^{D} (x_{t, i} - x_{t, i}^{(r)})^2, \\
    \mathcal{L}_{kld}
    &=\frac{1}{2T} \sum_{t=-T+1}^{0}\sum_{i=1}^{D_z}\left(1+\log \sigma_{t,i}^{2}-\mu_{t,i}^{2}-\sigma_{t,i}^{2}\right),
\end{aligned}
\end{equation}
where the MSE loss function is utilized for $\mathcal{L}_{rec}$ and $x_t^{(r)}$ is the reconstructed version of feature $x_t$. Note that the feature $x_t$ is from viewpoint $v_k$ if multi-view setting is applied (\emph{e.g.}, $v_{12}$-$v_3$).

The three losses are combined to form the comprehensive loss function of the network:

\begin{equation}
\label{eq:final_loss}
\mathcal{L} = \lambda_1 \mathcal{L}_{cls} + \lambda_2 \mathcal{L}_{rec} + \lambda_3 \mathcal{L}_{kld}.
\end{equation}

In this equation, the $\lambda_1$, $\lambda_2$, and $\lambda_3$ are trade-off weights. 
In table IX\begin{algorithm}[htbp]
    \caption{PTMA Training paradigm}
    \hypertarget{alg:ptma_training_alg}
    \small
    \SetAlgoLined
    \SetKwInOut{Input}{Input}
    \SetKwInOut{Output}{Output}
    
    \Input{Feature sequence $\mathbf{X}^{v_i}=\{x_t^{v_i}\}_{t=-T+1}^{t=0}$, Corresponding GRU $\mathbf{\Phi}_{gru}$ and TMA $\mathbf{\Phi}_{tma}$ in GRU-TMA Cell, Linear classifier $\mathbf{\Phi}_{c}$, Probabilistic encoder $\mathbf{\Phi}_{pe}$ and decoder $\mathbf{\Phi}_{pd}$, Expansion layer $f_q$.}

    \Output{Final loss $\mathcal{L}$.}
    Initialize $\mathbf{\Phi}_{gru}$, $\mathbf{\Phi}_{tma}$, $\mathbf{\Phi}_{c}$, $\mathbf{\Phi}_{pe}$, $\mathbf{\Phi}_{pd}$, $f_q$\;

    \For{$E$ epochs}{
        \For{$t=1$ \bfseries to T}{

            \textcolor{gray}{\# Probabilistic branch}\\
            Compute $\boldsymbol{\mu}$: ${\mu}_t \leftarrow \text{MLP}_1(\mathbf{\Phi}_{pe}(x^{v_i}_t))$\;

            Compute $\boldsymbol{\sigma}$: ${\sigma}_t \leftarrow \text{MLP}_2(\mathbf{\Phi}_{pe}(x^{v_i}_t))$\;

            Sample $\bm{\varepsilon}$: ${\varepsilon}_t \sim \mathcal{N}(\mathbf{0}, \mathbf{I})$\;
            Compute $\mathbf{z}$: ${z}_t \leftarrow \text{Reparameterized}(\bm{\varepsilon}_t, {\mu}_t, {\sigma}_t)$\;
            Compute $\mathbf{X}^{(r)}$: $x^{v_i(r)}_t \leftarrow \text{MLP}({z}_t)$\;\
            
            \textcolor{gray}{\# Classification branch}\\
            Compute $h$: $h_t \leftarrow \mathbf{\Phi}_{gru}(x^{v_i}_t, h_{t-1})$\;
            Update $\mathbf{H}$: $\mathbf{H}_{:t} \leftarrow [\mathbf{H}_{:t-1}|h_t]$\;\

            \textcolor{gray}{\# Refined encoding from TMA using Eq. \ref{eq:temporal_mask}, Eq. \ref{eq:tma}}\\

            Compute $\mathcal{M}$\;
            Compute $\tilde{\mathbf{H}}$: $\tilde{{h}}_{t} \leftarrow \mathbf{\Phi}_{tma}(\mathbf{H}_{:t}, f_q({z}_t), \mathcal{M})$\;\

            \textcolor{gray}{\# Compute losses}\\
            Compute $\hat{\mathbf{y}}$: $\hat{y}_t \leftarrow \mathbf{\Phi}_{c}(\tilde{{h}}_{t})$\;

            Update $\mathcal{L}_{cls},\ \mathcal{L}_{rec},\ \mathcal{L}_{kld}$ using Eq. \ref{eq:cls}, Eq. \ref{eq:kld_rec}\;
        }
        Compute $\mathcal{L}$ using Eq. \ref{eq:final_loss}: $\mathcal{L} \leftarrow \lambda_1 \mathcal{L}_{cls} + \lambda_2 \mathcal{L}_{rec} + \lambda_3 \mathcal{L}_{kld}$.
    }
\end{algorithm}

\section{Experiments}
\label{sec:experiments}

In this section, we validate the effectiveness of the proposed PTMA method on the DAHLIA, IKEA ASM, and Breakfast datasets, applying three evaluation mechanisms. Additionally, various ablation experiments are conducted to further elucidate the impact of each component of the method. All OAD models are implemented using the official code, with careful fine-tuning applied (\emph{e.g.}, different learning rates, length of long and short memories, epochs, embedding dimensions).

\begin{table*}[ht]
    \setlength{\tabcolsep}{1.5 mm}
    \centering
    \small

    \captionsetup{aboveskip=3pt}
    \caption{Comparative experiments of proposed PTMA model with OAD models on DAHLIA, IKEA ASM, and Breakfast dataset under cs, cv, and csv evaluation mechanisms.}
    \label{tab:comparative}
    \vskip 0.0in
    \renewcommand{\arraystretch}{1.2}  
    \begin{tabular}{llcccccccccc}
    \toprule[1.2pt] 
    \multirow{2}*{Datasets} & \multirow{2}*{Method} & \multicolumn{4}{c}{cs} & \multicolumn{3}{c}{cv}  &  \multicolumn{3}{c}{csv}    \\
    \cmidrule(lr){3-6} \cmidrule(lr){7-9} \cmidrule(lr){10-12}
    ~  & ~     & $v_1$  & $v_2$ & $v_3$ & Avg. & $v_1$-$v_2$ & $v_1$-$v_3$ & Avg. & $v_1$-$v_2$ & $v_1$-$v_3$ & Avg.\\ \hline 

    \multirow{5}*{DAHLIA} & OadTR \cite{Wang_2021_OadTR} & 61.71 & 53.41 & 59.19 & 58.10 & 30.16 & 31.14 & 30.65 & 31.58 & 27.05 & 29.32 \\
    ~ & LSTR \cite{Xu_2021_Long} & 77.78 & 67.74 & 73.57 & 73.03 & 37.17 & 47.56 & 42.36 & 39.26 & 41.46 & 40.36 \\
    ~ & MAT \cite{Wang_2023_Memory} & 72.89 & 71.92 & 75.11 & 73.31 & 24.32 & 36.96 & 30.64 & 26.90 & 30.77 & 28.84 \\
    ~ & MiniROAD \cite{An_2023_MiniROAD} & 71.47 & 65.89 & 71.77 & 69.71 & 33.11 & 38.04 & 35.58 & 39.82 & 35.41 & 37.61 \\
    \rowcolor{light}
    ~ & \textbf{PTMA} & \textbf{85.36} & \textbf{76.86} & \textbf{82.01} & \textbf{81.41} & \textbf{43.45} & \textbf{52.07} & \textbf{47.76} & \textbf{45.28} & \textbf{41.96} & \textbf{43.62} \\
    \hline

    \multirow{5}*{IKEA ASM} & OadTR \cite{Wang_2021_OadTR} & 66.14 & 65.23 & 62.30 & 64.56 & 70.98 & 68.29 & 69.64 & 70.19 & 65.61 & 67.90 \\
    ~ & LSTR \cite{Xu_2021_Long} & 67.50 & 67.27 & 60.23 & 65.00 & 71.17 & 70.84 & 71.00 & 70.43 & 64.57 & 67.50 \\
    ~ & MAT \cite{Wang_2023_Memory} & 77.80 & 74.60 & 67.27 & 73.22 & 82.19 & 79.29 & 80.74 & 74.99 & 69.11 & 72.05 \\
    ~ & MiniROAD \cite{An_2023_MiniROAD} & 86.15 & 82.57 & 71.10 & 79.94 & 89.41 & 87.60 & 88.50 & 83.59 & 76.39 & 79.99 \\
    \rowcolor{light}
    ~ & \textbf{PTMA} & \textbf{86.99} & \textbf{86.72} & \textbf{75.77} & \textbf{83.16} & \textbf{92.48} & \textbf{90.03} & \textbf{91.26} & \textbf{87.26} & \textbf{81.68} & \textbf{84.47} \\
    \hline

    \multirow{5}*{Breakfast} & OadTR \cite{Wang_2021_OadTR} & 64.16 & \textbf{75.05} & 67.30 & 68.84 & 57.67 & 60.79 & 59.23 & 56.90 & 57.88 & 57.39 \\
    ~ & LSTR \cite{Xu_2021_Long} & 62.74 & 68.03 & 67.10 & 65.96 & 60.26 & 60.77 & 60.52 & 54.97 & 58.30 & 56.64 \\
    ~ & MAT \cite{Wang_2023_Memory} & 62.42 & 69.27 & 67.65 & 66.45 & 59.06 & 59.16 & 59.11 & 54.08 & \textbf{59.11} & 56.60 \\
    ~ & MiniROAD \cite{An_2023_MiniROAD} & 61.62 & 71.74 & 64.94 & 66.10 & 58.38 & 55.21 & 56.80 & 54.92 & 54.18 & 54.55 \\
    \rowcolor{light}
    ~ & \textbf{PTMA} & \textbf{66.06} & 74.49 & \textbf{71.55} & \textbf{70.70} & \textbf{63.49} & \textbf{61.17} & \textbf{62.33} & \textbf{61.36} & 57.18 & \textbf{59.27}\\

    \bottomrule[1.2pt]

    \end{tabular}
    

\end{table*}
\subsection{Datasets and setup}
\label{sec:data_setup}
\textbf{Datasets. IKEA ASM} dataset \cite{BenShabat_2021_IKEA} offers 371 assembly sequences across four furniture types, captured from three views, for a total of 1,113 videos. It's a rich resource for studying assembly actions with an average of 23 segmented actions per video and 32 distinct action classes. The \textbf{DAHLIA} dataset \cite{Vaquette_2017_DAily}, using three Kinect v2 sensors, presents unscripted activities over 40 minutes per participant, focusing on 7 high-level action classes. The \textbf{Breakfast} dataset, as featured in \cite{Kuehne_2014_Language}, contains 1.7k cooking videos of 10 dishes, averaging 6.9 action segments each, recorded over 77 hours with diverse camera setups. We sample the shared action classes in all four splits from \cite{Kuehne_2014_Language} and obtain 38 actions at last. Due to missing camera views, experiments focus on view ``cam01'', ``webcam01'', and ``webcam02''.

\textbf{Feature Extraction.}
\label{sec:feature_extraction}
For extracting features from IKEA ASM and DAHLIA videos, we adhere to the methods established in prior OAD researches \cite{Xu_2021_Long, Wang_2021_OadTR}, utilizing a dual-branch network architecture known as TSN \cite{Wang_2016_Temporal}. A ResNet50 pre-trained on the Kinetics dataset \cite{Carreira_2017_Quo} is applied as the backbone with dimensions of 4096 (RGB 2048 + Flow 2048). For the DAHLIA dataset, videos are processed at a frame rate of 15 FPS with a segment length of 5 frames, while the videos from IKEA ASM are encoded at a frame rate of 25 FPS with the same segment length. The extraction process utilizes an open-source toolkit \cite{Contributors_2020_OpenMMLabs}. For the Breakfast dataset, we adopt I3D features pre-extracted in \cite{Kuehne_2014_Language}, with dimensions of 2048.

\textbf{Implementation Details.}
We conduct all comparative and ablation experiments using NVIDIA Tesla V100 GPUs. The training is performed with the Adam optimizer \cite{Kingma_2014_Adam}, utilizing a learning rate with cosine annealing \cite{Loshchilov_2016_Sgdr}, fixed at an initial value of 0.00014. The temporal window lengths for the DAHLIA, IKEA ASM, and Breakfast datasets are set to 512, 256, and 512, respectively. For the compressed latent variable $\mathbf{z}$, we specify a fixed dimension of 256, 768, and 512 for DAHLIA, IKEA ASM, and Breakfast. During the training phase, epochs are set as 10, and batch size is 16. An early stopping trick is employed in the training.

\begin{figure}[htbp]
    \centering
    \includegraphics[width=0.45\textwidth]{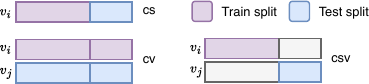}
    \caption{Illustration of three evaluation mechanisms. $v_i$ and $v_j$ are different viewpoints.}
    \label{fig:evaluation}
  \end{figure}

\textbf{Evaluation Mechanism.}
To accommodate the cross-view setup of OAD tasks, in line with multi-view action recognition works \cite{Zhang_2013_Cross, Ullah_2021_Conflux, Das_2020_VPN}, we establish three evaluation mechanisms for assessing cross-viewpoint OAD models: cross-subject (\textbf{cs}), cross-view (\textbf{cv}), and cross-subject-view (\textbf{csv}) as depicted in Figure \ref{fig:evaluation}. cs is used to evaluate the OAD capability of the model with a fixed viewpoint across different experimental subjects. cv assesses the model's generalization ability to the same subject across different viewpoints. csv challenges the model by evaluating both aspects simultaneously, making it a very demanding test mechanism.

\textbf{Metrics.}
Similar to previous works \cite{Xu_2021_Long, Wang_2021_OadTR}, the per-frame mean Average Precision (mAP) is applied to DAHLIA, and mean calibrated Average Precision (mcAP) \cite{DeGeest_2016_Online} to IKEA ASM and Breakfast due to their imbalanced distribution of positive and negative samples. defined as Eq. \ref{eq:mcAP}. 

\begin{equation}
\label{eq:mcAP}
\begin{aligned}
cAP =\frac{\sum_{k} c \operatorname{Prec}(k) * I(k)}{P},
\end{aligned}
\end{equation}
where $cPrec=\frac{w * T P}{w * T P+F P}$, $I(k)=1$ if the $k$-th frame is a True Positive (TP), $P$ counts the number of TPs and $w$ is the ratio between the negatives and positives.

\subsection{Comparative experiments}

We conduct comparative experiments with the PTMA model on the DAHLIA, IKEA ASM and Breakfast datasets under cs, cv, and csv evaluation mechanisms. The Avg. means the average level of experiments across the shared evaluation settings. The PTMA model is compared with several OAD models from recent years, with the results presented in Table \ref{tab:comparative}. The results indicate that under the cs setup, PTMA outperforms other OAD models across all three datasets in terms of mAP or mcAP except for a slight disadvantage on the Breakfast dataset (\emph{e.g.}, 74.49\%, cs, $v_2$). In the cv and csv setups, our model maintains considerable competitiveness and achieves significant leadership on the DAHLIA and IKEA ASM datasets (\emph{e.g.}, 91.26\% and 84.47\% on IKEA ASM). 

According to the results, we observe that even though the probabilistic branch only incorporates a single viewpoint during training, the PTMA model can still effectively generalize to unseen viewpoints. This indicates that the area $ S_1 $ is in close proximity to the area $ S $ as depicted in Figure \ref{fig:latent_space} (\emph{i.e.}, $S_1 / S \rightarrow 1$), thereby confirming the PTMA's robustness in cross-view OAD tasks.

\subsection{Ablation studies}

\begin{table}[htbp]
    \caption{Ablation study on PTMA model with multi-view source for training while unseen view for testing using mcAP(\%) on DAHLIA and IKEA ASM dataset.}
    \belowrulesep=0pt
    \aboverulesep=0pt
    \centering
    \small

    \setlength{\tabcolsep}{0.6 mm}
    \renewcommand{\arraystretch}{1.2} 

    \begin{tabular}{l|cccccc}
        \toprule[1.2pt]
        \multirow{2}*{\diagbox[width=8em,trim=r,linewidth=0.2pt]{Datasets}{m-c(s)v}} & \multicolumn{3}{c}{m-cv} & \multicolumn{3}{c}{m-csv} \\
        \cmidrule(lr){2-4}\cmidrule(lr){5-7}
        ~ & $v_{13}$-$v_2$ & $v_{12}$-$v_3$ & Avg. & $v_{13}$-$v_2$ & $v_{12}$-$v_3$ &Avg.  \\ \hline
        DAHLIA & 44.74 & 52.47 & \textbf{48.61} & 47.11 & 44.22 & \textbf{45.67} \\
        IKEA ASM & 92.58 & 90.53 & \textbf{91.56} & 87.49 & 81.76 & \textbf{84.63} \\
        \bottomrule[1.2pt]
    \end{tabular}
    \label{tab:ablation_mcv}
\end{table}

\textbf{Multi-view training with cross-view testing.} To further effectively leverage the multi-view data, we engage multiple view-level sources in training (\emph{i.e.}, m-c(s)v setting), aiming to learn a distribution that is richer in view-invariant properties than the underlying distribution itself. In Table \ref{tab:ablation_mcv}, $v_{ij}$-$v_k$ denotes using $v_i$ as the input for the classification branch and the probabilistic branch, employing $v_j$ as the reconstruction label for the probabilistic decoder, and applying view $v_k$ for testing (\emph{e.g.}, $v_{13}$-$v_2$). The results indicate that training with cross-view assistance in the probabilistic model can effectively enhance the cross-view OAD performance. On average, improvements of mAP or mcAP are made on the DAHLIA dataset by 3.3\%, and on the IKEA ASM dataset by 0.46\%. 

While multi-view sources greatly enhance cross-view OAD performance, they impose higher demands on the synchronization of input data. Experiments in Table \ref{tab:comparative} have demonstrated that probabilistic models trained from a single viewpoint are capable of compressing video features into lower-dimensional latent encodings that can adapt to videos from previously unseen viewpoints. Consequently, the additional ablation studies primarily focus on settings with single-view training.

\begin{table}[htbp]
    \centering
    \small

    \caption{Ablation study on different parts of probabilistic modeling using mcAP(\%) on IKEA ASM dataset.}

    \setlength{\tabcolsep}{1.2 mm}
    \label{tab:ablation_ptma}
    \renewcommand{\arraystretch}{1.2}  
        \begin{tabular}{lcccccc}
            \toprule[1.2pt]
            \multirow{2} * {$\mathcal{L}_{cls}$} & \multirow{2} * {$\mathcal{L}_{rec}$} & \multirow{2} * {$\mathcal{L}_{kld}$} & \multicolumn{2}{c}{cv} & \multicolumn{2}{c}{csv} \\
            \cmidrule(lr){4-5} \cmidrule(lr){6-7}
            ~ & ~ & ~ & $v_1$-$v_2$ & $v_1$-$v_3$ & $v_1$-$v_2$ & $v_1$-$v_3$ \\ \hline
            $\checkmark$ & $\bf{-}$ & $\bf{-}$ & 89.18 & 88.60 & 84.32 & 79.51\\
            $\checkmark$ & $\times$ & $\times$ & 92.05 & 88.95 & 84.59 & 77.06 \\ 
            $\checkmark$ & $\checkmark$ & $\times$ & 92.28 & 89.70 & 86.95 & 81.54 \\ 
            $\checkmark$ & $\times$ & $\checkmark$ & 91.45 & 89.24 & 86.03 & 80.90 \\
            \rowcolor{light} 
            $\checkmark$ & $\checkmark$ & $\checkmark$ & \textbf{92.48} & \textbf{90.03} & \textbf{87.26} & \textbf{81.68} \\ 

            \bottomrule[1.2pt]

        \end{tabular}
\end{table}
\textbf{Ablation studies of probabilistic model.}
In the PTMA model, the reconstruction loss ensures that the compressed latent variable adequately represents the original feature input, while the KLD term guarantees that this variable is robust, closely approximating the true distribution $p(\mathbf{z})$. We have conducted ablation studies on these two components of the loss within the probabilistic model, with the findings detailed in Table \ref{tab:ablation_ptma}. The ``$\checkmark$ $\bf{-}$ $\bf{-}$'' in the first row represents the performance of using a carefully tuned GRU model as a baseline for comparison. It's noteworthy that neglecting the KLD loss reduces the probabilistic model to a standard autoencoder. Without probabilistic modeling, the model's average mcAP on the IKEA ASM dataset is 88.89\% (cv) and 81.91\% (csv). This is lower than when probabilistic modeling is used, which achieves 90.35\% for cv and 83.47\% for csv. Thus, while non-probabilistic modeling can adequately represent view-level features, it does not perform as well as probabilistic modeling.

Furthermore, disregarding both the reconstruction loss and the KLD loss transforms the probabilistic model into a learnable query mechanism. All the training modes are depicted in Figure \ref{fig:overview}. Incorporating both the reconstruction loss and KLD loss, PTMA achieves optimal results, averaging an mcAP of 87.86\%. This balance is crucial for harnessing the potential of the model in cross-view OAD tasks.

\begin{table}[htbp]
    \centering
    \small

    \caption{Ablation study on latent dimensions $D_z$ of probabilistic model, tested using mcAP(\%) on IKEA ASM dataset.}

    \centering
    \setlength{\tabcolsep}{1.2 mm}
    \label{tab:ablation_latent}
    \renewcommand{\arraystretch}{1.2}  
    \begin{tabular}{lcccc}
    \toprule[1.2pt]
    \multirow{2} * {$D_z$} & \multicolumn{2}{c}{cv} & \multicolumn{2}{c}{csv} \\
    \cmidrule(lr){2-3} \cmidrule(lr){4-5}
    ~ & $v_1$-$v_2$ & $v_1$-$v_3$ & $v_1$-$v_2$ & $v_1$-$v_3$ \\ \hline

    256 & 91.83 & 89.99 & 86.77 & 81.38 \\
    512 & 91.49 & 89.60 & 86.65 & 80.33 \\
    768 & \textbf{92.48} & \textbf{90.03} & 86.71 & 80.66 \\
    1024 & 91.88 & 89.58 & \textbf{87.26} & \textbf{81.68} \\
    \bottomrule[1.2pt]

    \end{tabular}

\end{table}

\textbf{Different latent dimensions.}
The dimensionality of the latent variable $\mathbf{z}$ determines the complexity of the latent compressed information. A high dimensionality may introduce noise in view-level features, whereas a low dimensionality risks omitting critical action features. We have conducted experiments with different dimensions on the IKEA ASM dataset, and the results are presented in Table \ref{tab:ablation_latent}. The results reveal that overly compressed representations (\emph{e.g.}, $D_z=256$), can fail to capture the full view-invariant information. Conversely, a dimensionality that is too high (\emph{e.g.}, $D_z=1024$), may incorporate excessive noise under cv setting. Optimal results for the IKEA ASM dataset are achieved with $D_z=768$ for cv and $D_z=1024$ for csv, striking a balance that enhances the model's ability to generalize across views without being confounded by noise or losing essential features.

\begin{table}[htbp]
    \caption{Ablation study on PTMA model with different temporal window sizes using mcAP(\%) on DAHLIA and IKEA ASM dataset.}

    \centering
    \small
    \setlength{\tabcolsep}{1.2 mm}
    \label{tab:ablation_ws}
    \renewcommand{\arraystretch}{1.2}  
    \begin{tabular}{lccccc}
        \toprule[1.2pt]
        \multirow{2} * {Dataset} & \multirow{2} * {$T$} & \multicolumn{2}{c}{cv} & \multicolumn{2}{c}{csv} \\
        \cmidrule(lr){3-4} \cmidrule(lr){5-6}
        ~ & ~ & $v_1$-$v_2$ & $v_1$-$v_3$ & $v_1$-$v_2$ & $v_1$-$v_3$ \\ \hline
    
        \multirow{3}*{DAHLIA} & 128 & 37.16 & 43.74 & 37.74 & 39.86 \\
        ~ & 256 & 42.25 & 50.68 & 37.92 & 36.77 \\
        ~ & 512 & \textbf{43.45} & \textbf{52.07} & \textbf{45.28} & \textbf{41.96} \\ \hline
        
        \multirow{3}*{IKEA ASM} & 128 & 91.83 & 89.13 & 86.42 & 80.06 \\
        ~ & 256 & \textbf{92.48} & \textbf{90.03} & \textbf{86.71} & 80.66 \\
        ~ & 512 & 92.32 & 88.49 & 86.14 & \textbf{81.12} \\  
        \bottomrule[1.2pt]

    \end{tabular}

\end{table}

\textbf{Different temporal window size.}
During training, we utilize a temporal window of length $T$ to sample video sequences, which dictates the model's ``field of view'' for feature observation. Experiments with $T$ on the DAHLIA and IKEA ASM datasets, detailed in Table \ref{tab:ablation_ws}, indicate optimal settings of $T=256$ for IKEA ASM and $T=512$ for DAHLIA. These choices reflect the average action durations, with IKEA ASM favoring a shorter window to avoid dense action interference, while DAHLIA accommodates a longer one due to less action variety. Consequently, in cross-view scenarios, the PTMA model achieves an average best performance of 45.69\% (mAP) on DAHLIA and 87.86\% (mcAP) on IKEA ASM, respectively.

\begin{figure}[htbp]
    \centering
    \subfloat[]{
        \includegraphics[width=0.15\textwidth]{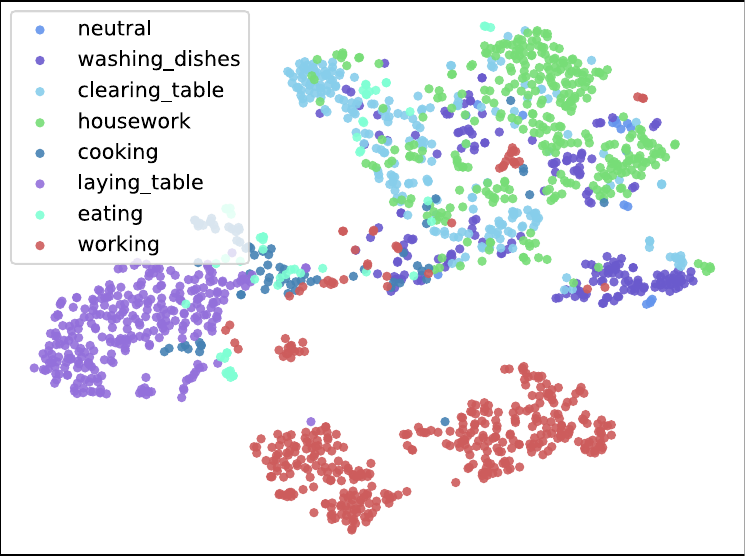}}
    \label{fig:tsne-K1}
    \hfill
    \subfloat[]{
        \includegraphics[width=0.15\textwidth]{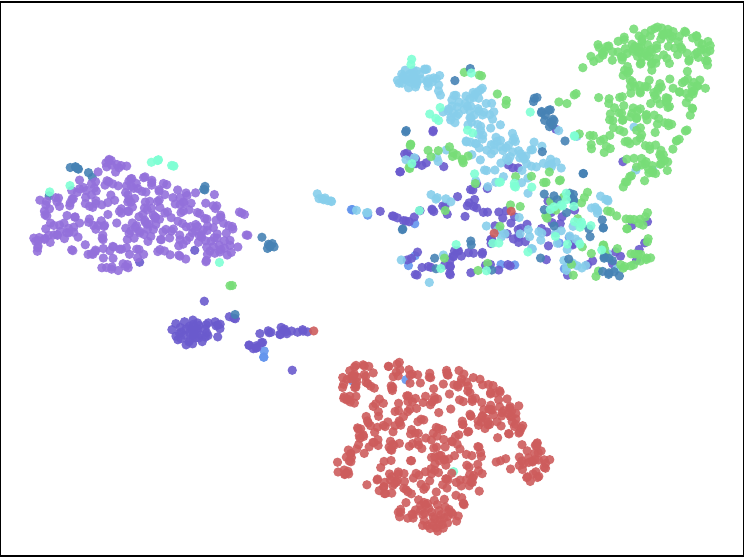}}
        \label{fig:tsne-K2}
    \hfill
    \subfloat[]{
        \includegraphics[width=0.15\textwidth]{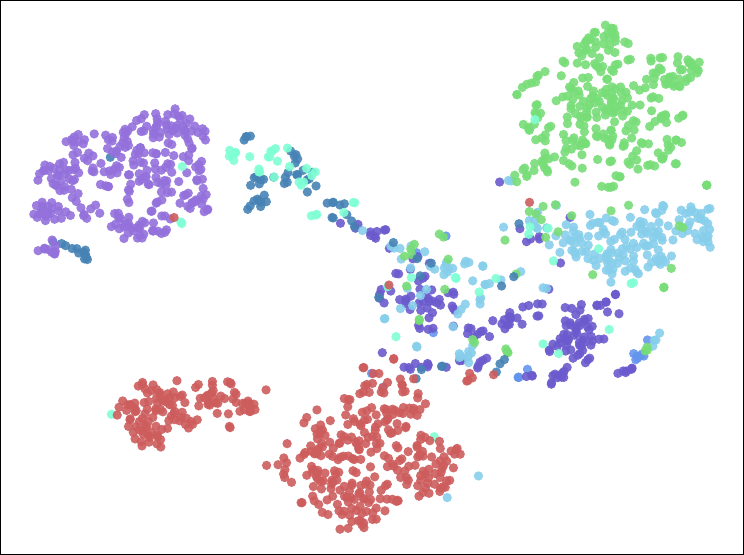}}
    \label{fig:tsne-K3}
    \caption{The t-SNE visualization of latent encodings from the probabilistic encoder. (a), (b), and (c) refer to the latent encodings of view $v_1$, $v_2$, and $v_3$.}
    \label{fig:tsne}
\end{figure}

\textbf{Visulization of latent encodings.}
In Figure \ref{fig:tsne}, we present the t-SNE encodings of the latent vectors output by the probabilistic model trained on the DAHLIA dataset during the inference for videos of view $v_1$, $v_2$, and $v_3$, respectively. Despite the probabilistic encoder being trained on view $v_1$, its encoding capability remains robust on view $v_2$ and $v_3$. The encodings of different actions are separated and the same actions are distributed closely in high-dimensional space, indicating that the probabilistic model can learn refined latent view-level information common across different views under a cross-view setting. It is observed that for action samples with a high frequency of occurrence, such as working, features from both view $v_2$ and $v_3$ are effectively extracted. However, discrepancies are noted within the feature encodings between view $v_1$ and view $v_3$, but not in view $v_2$. Thus, it is concluded that this method exhibits some overfitting issues. Meanwhile, in samples with fewer occurrences, such as washing dishes, the feature encoding distribution is relatively scattered, indicating a potential underfitting problem. Therefore, although the proposed PTMA model has demonstrated good generalization in unseen viewpoints, issues of overfitting with multiple samples and underfitting with few samples persist, largely depending on the number of samples. Addressing this issue will be one of our primary focuses for future work. More t-SNE visualizations of the latent vectors are provided in the supplementary materials.

\begin{table}[htbp]
    \centering
    \small
    \setlength{\tabcolsep}{1. mm}
    \caption{The categories with the best and worst action detection results on the IKEA ASM dataset. ``proportion\%'' represents the percentage of each action category in terms of time.}
    \label{tab:best2worst}
    \renewcommand{\arraystretch}{1.2}  
    \begin{tabular}{lcc}
        \toprule[1.2pt]
        Action class & Proportion (\%) & mcAP (\%)\\ \hline

        attach shelf to table  & 7.10 & 96.65 \\ 
        pick up pin  & 0.52 & 95.61 \\ 
        pick up bottom panel  & 0.52 & 94.85 \\ 
        lay down bottom panel  & 0.14 & 94.65 \\ 
        slide bottom of drawer  & 1.81 & 94.20 \\ 
        position the drawer right side up & 0.53 & 94.17 \\ 
        ...... \\
        pick up front panel  & 0.16 & 83.14 \\
        pick up table top  & 0.11 &  79.77\\ 
        lay down leg  & 0.07 & 76.81 \\ 
        lay down back panel  & 0.03 & 74.91 \\ 
        lay down shelf  & 0.11 & 73.90 \\ 
        push table  & 0.06 & 64.49 \\

        \bottomrule[1.2pt]

    \end{tabular}
\end{table}
\textbf{Well performed and poorly performed categories.} 
We investigate the reasons behind the performance of well-performed and poorly-performed samples and determine that the model's performance is highly correlated with the abundance of data for various action categories. The action categories that perform best and worst in the Ikea ASM dataset, along with respective proportions within the dataset, are demonstrated in Table \ref{tab:best2worst}. The findings indicate that action categories with a larger scale of the data tend to have better performance in cross-view OAD tasks.

\begin{table}[htbp]
    \centering
    \small
    \setlength{\tabcolsep}{1. mm}
    \caption{Ablation study on OAD models in terms of parameter scale, GFLOPs, inference speed, and mcAP(\%) on IKEA ASM dataset.}
    \label{tab:ablation_params}
    \renewcommand{\arraystretch}{1.2}  
    \begin{tabular}{lcccc}
        \toprule[1.2pt]
        Method & Params & GFLOPs & FPS & mcAP (\%)\\ \hline

        OadTR \cite{Wang_2021_OadTR} & 75.8M & 2.55 & 110 & 67.90 \\ 
        LSTR \cite{Xu_2021_Long} & 58.0M & 7.53 & 92 & 67.50 \\ 
        MAT \cite{Wang_2023_Memory} & 94.6M & 20.2 & 73 & 72.05 \\ 
        MiniROAD \cite{An_2023_MiniROAD} & \textbf{10.5M} & \textbf{1.08} & \textbf{32510} & 79.99 \\ 
        PTMA  & 12.9M & 1.96 & 29110 & \textbf{84.47} \\ 
        \bottomrule[1.2pt]

    \end{tabular}
\end{table}
\textbf{Parameters count and Inference performance.}
In Table \ref{tab:ablation_params}, we compare the PTMA's parameter count, GFLOPs, and inference speed with traditional OAD models in the csv setting on the IKEA ASM dataset using mcAP. It reveals that while our model achieves the highest performance (84.47\%), it has a slightly higher parameter count and GFLOPs compared to MiniROAD \cite{An_2023_MiniROAD} (\emph{e.g.}, 12.9M vs. 10.5M, 1.96G vs. 1.08G). Nonetheless, the PTMA model's inference speed and overall efficiency meet the practical requirements, making it a competitive model for cross-view OAD tasks.

\begin{table}[htbp]
    \small
    \caption{Compression study on PTMA with different parameter scale.}
    \label{tab:compression}
    \centering
    \renewcommand{\arraystretch}{1.2}  
    \begin{tabular}{l|cccc}
    \toprule[1.2pt]

    Parmas & 12.9M & 10.6M & 8.7M & 7.2M \\ 
    \hline
    mcAP (\%) & 87.26 & 86.71 & 86.65 & 86.77 \\

    \bottomrule[1.2pt]
    \end{tabular}
    
\end{table}

\textbf{Different parameter scales.}
The dimension of the probabilistic encoder's output is decreased to control the parameter scales of PTMA model. According to the results researched on IKEA ASM dataset in Table \ref{tab:compression}, although the accuracy performance on the dataset only shows slight fluctuations with the decrease of model parameter scale, it still remains within an acceptable range. This phenomenon suggests that the proposed PTMA model can maintain a normal operating level in resource constrained environments.

\begin{table}[htbp]
    \small
    \centering
    \caption{Ablation study on PTMA with mixed viewpoints.}
    \label{tab:adapt_view}
    \renewcommand{\arraystretch}{1.2}  
    \begin{tabular}{l|ccccc}
    \toprule[1.2pt]
    
    Method & $v_{1}-v_{23}$ & $v_{2}-v_{13}$ & $v_{3}-v_{12}$ & Avg.\\
    \hline
    MiniROAD \cite{An_2023_MiniROAD} & 80.41 & 80.84 & 75.68 & 78.98 \\ 
    MAT \cite{Wang_2023_Memory} & 71.60 & 73.94 & 74.13 & 73.22 \\ 

    PTMA & 83.51 & 84.32 & 84.40 & 84.08\\ 

    \bottomrule[1.2pt]
    \end{tabular}
    
\end{table}

\textbf{Mixed testing viewpoints.}
    In Table \ref{tab:adapt_view}, the model's inference is expanded by training with a specific viewpoint and testing with mixed viewpoints. For example, the setting ``$v_{1}$-$v_{23}$'' indicates training with view $v_1$ and testing with mixed views $v_2$ and $v_3$. The average mcAP of MiniROAD \cite{An_2023_MiniROAD} and MAT \cite{Wang_2023_Memory} are 78.98\% and 84.08\%, which is far lower than that of the PTMA model (84.08\%). It illustrates that the proposed PTMA model maintains a high level of robustness across different testing view settings, thereby accounting for its performance.

\section{Limitations}  
\textbf{Trade-off between real-time performance and computational efficiency:} Although the module is designed to optimize efficiency, the model's real-time performance could still be constrained when handling multiple concurrent action instances. The computational overhead associated with feature extraction and sequential modeling may lead to latency, especially when deployed on resource-limited edge devices.

\textbf{Limited adaptability to extreme viewpoint variations:} While PTMA performs well under common viewpoint variations, its generalization might be limited in extreme scenarios, such as high overhead, low ground-level, or oblique viewpoints. These viewpoints may introduce severe distortions, occlusions, or motion blur, making it challenging for the model to extract discriminative spatial-temporal features effectively. Additional adaptation strategies or viewpoint-invariant representations may be needed to improve robustness.

\textbf{Challenges in long-tail action detection:} The model's detection performance may degrade for low-frequency actions in cases of imbalanced action category distributions. Rare actions often lack sufficient training samples, leading to biased predictions toward more frequent actions. This imbalance can result in lower recall and precision for underrepresented classes. Addressing this issue may require advanced data augmentation techniques, class rebalancing strategies, or meta-learning approaches to enhance generalization for long-tail action categories.

\section{Conclusion}
In this paper, we introduce the PTMA model to address the challenges of viewpoint variability in OAD tasks. The PTMA leverages probabilistic modeling to generate latent compressed representations, which are integrated with a GRU-TMA cell, performing temporal masked attention for robust autoregressive frame-level video analysis. If multi-view sources are permitted, the probabilistic modeling component of the PTMA can be enhanced by reconstructing the original video data from various viewpoints, which aids in the extraction of features that are invariant to the viewing perspective. Through extensive experiments across three evaluation mechanisms—cross-subject, cross-view, and cross-subject-view—we demonstrate the superior generalization capabilities of the PTMA model. Our method achieves state-of-the-art performance on the DAHLIA, IKEA ASM, and Breakfast datasets, underscoring its effectiveness in mitigating the limitations of traditional OAD models.

Future work will focus on enhancing the encodings of the PTMA model for view-invariant features under single-view training conditions. This advancement aims to improve the model's performance in complex multi-view scenarios, particularly for cross-view online action detection tasks.

\bibliography{ref}

\begin{thebibliography}{10}
\providecommand{\url}[1]{#1}
\csname url@samestyle\endcsname
\providecommand{\newblock}{\relax}
\providecommand{\bibinfo}[2]{#2}
\providecommand{\BIBentrySTDinterwordspacing}{\spaceskip=0pt\relax}
\providecommand{\BIBentryALTinterwordstretchfactor}{4}
\providecommand{\BIBentryALTinterwordspacing}{\spaceskip=\fontdimen2\font plus
\BIBentryALTinterwordstretchfactor\fontdimen3\font minus \fontdimen4\font\relax}
\providecommand{\BIBforeignlanguage}[2]{{%
\expandafter\ifx\csname l@#1\endcsname\relax
\typeout{** WARNING: IEEEtran.bst: No hyphenation pattern has been}%
\typeout{** loaded for the language `#1'. Using the pattern for}%
\typeout{** the default language instead.}%
\else
\language=\csname l@#1\endcsname
\fi
#2}}
\providecommand{\BIBdecl}{\relax}
\BIBdecl

\bibitem{Pavlidis_2001_Urban}
I.~Pavlidis, V.~Morellas, P.~Tsiamyrtzis, and S.~Harp, ``Urban surveillance systems: from the laboratory to the commercial world,'' \emph{Proc. IEEE}, vol.~89, no.~10, pp. 1478--1497, 2001.

\bibitem{Liu_2022_EHPE}
H.~Liu, T.~Liu, Y.~Chen, Z.~Zhang, and Y.-F. Li, ``Ehpe: Skeleton cues-based gaussian coordinate encoding for efficient human pose estimation,'' \emph{IEEE Trans. Multimedia}, 2022.

\bibitem{Xie_2021_Graph}
L.~Xie, Y.~Luo, S.-F. Su, and H.~Wei, ``Graph regularized structured output svm for early expression detection with online extension,'' \emph{IEEE Trans. Cybern.}, vol.~53, pp. 1419--1431, 2021.

\bibitem{Fang_2023_HODN}
S.~Fang, Z.~Lin, K.~Yan, J.~Li, X.~Lin, and R.~Ji, ``Hodn: Disentangling human-object feature for hoi detection,'' \emph{IEEE Trans. Multimedia}, 2023.

\bibitem{Liu_2018_Imitation}
Y.~Liu, A.~Gupta, P.~Abbeel, and S.~Levine, ``Imitation from observation: Learning to imitate behaviors from raw video via context translation,'' in \emph{Proc. IEEE Int. Conf. Robot. Autom.}, 2018, pp. 1118--1125.

\bibitem{Yu_2020_Spatio}
C.~Yu, X.~Ma, J.~Ren, H.~Zhao, and S.~Yi, ``Spatio-temporal graph transformer networks for pedestrian trajectory prediction,'' in \emph{Proc. Eur. Conf. Comput. Vis.}, 2020, pp. 507--523.

\bibitem{Liu_2023_Petrv2}
Y.~Liu, J.~Yan, F.~Jia, S.~Li, A.~Gao, T.~Wang, and X.~Zhang, ``Petrv2: A unified framework for 3d perception from multi-camera images,'' in \emph{Proc. IEEE Int. Conf. Comput. Vis.}, 2023, pp. 3262--3272.

\bibitem{Zhang_2024_Integration}
T.~Zhang, R.~Li, P.~Feng, and R.~Zhang, ``Integration of global and local knowledge for foreground enhancing in weakly supervised temporal action localization,'' \emph{IEEE Trans. Multimedia}, vol.~26, pp. 8476--8487, 2024.

\bibitem{Xia_2023_Exploring}
K.~Xia, L.~Wang, Y.~Shen, S.~Zhou, G.~Hua, and W.~Tang, ``Exploring action centers for temporal action localization,'' \emph{IEEE Trans. Multimedia}, vol.~25, pp. 9425--9436, 2023.

\bibitem{Ju_2022_Adaptive}
C.~Ju, P.~Zhao, S.~Chen, Y.~Zhang, X.~Zhang, Y.~Wang, and Q.~Tian, ``Adaptive mutual supervision for weakly-supervised temporal action localization,'' \emph{IEEE Trans. Multimedia}, vol.~25, pp. 6688--6701, 2022.

\bibitem{Zhu_2024_Part}
A.~Zhu, Q.~Ke, M.~Gong, and J.~Bailey, ``Part-aware unified representation of language and skeleton for zero-shot action recognition,'' in \emph{Proc. IEEE Conf. Comput. Vis. Pattern Recognit.}, 2024, pp. 18\,761--18\,770.

\bibitem{Wang_2024_CLIP}
X.~Wang, S.~Zhang, J.~Cen, C.~Gao, Y.~Zhang, D.~Zhao, and N.~Sang, ``Clip-guided prototype modulating for few-shot action recognition,'' \emph{Int. J. Comput. Vision}, vol. 132, no.~6, pp. 1899--1912, 2024.

\bibitem{Gan_2022_Temporal}
M.-G. Gan and Y.~Zhang, ``Temporal attention-pyramid pooling for temporal action detection,'' \emph{IEEE Trans. Multimedia}, vol.~25, pp. 3799--3810, 2022.

\bibitem{Tan_2024_Annealing}
Y.~Tan, L.~Xie, S.~Jing, S.~Fang, and K.~Zhang, ``Annealing temporal-spatial contrastive learning for multi-view online action detection,'' \emph{Knowl.-Based Syst.}, p. 112523, 2024.

\bibitem{Xie_2020_Efficient}
L.~Xie, W.~Guo, H.~Wei, Y.~Tang, and D.~Tao, ``Efficient unsupervised dimension reduction for streaming multiview data,'' \emph{IEEE Trans. Cybern.}, vol.~52, no.~3, pp. 1772--1784, 2020.

\bibitem{Leng_2023_Online}
H.~Leng, X.~Shi, W.~Zhou, K.~Zhang, Q.~Shi, and P.~Zhu, ``Online action detection with learning future representations by contrastive learning,'' in \emph{Proc. Int. Conf. Multimedia and Expo}, 2023, pp. 2213--2218.

\bibitem{Guo_2022_Uncertainty}
H.~Guo, H.~Wang, and Q.~Ji, ``Uncertainty-guided probabilistic transformer for complex action recognition,'' in \emph{Proc. IEEE Conf. Comput. Vis. Pattern Recognit.}, 2022, pp. 20\,052--20\,061.

\bibitem{Eun_2021_Temporal}
H.~Eun, J.~Moon, J.~Park, C.~Jung, and C.~Kim, ``Temporal filtering networks for online action detection,'' \emph{Pattern Recognit.}, vol. 111, p. 107695, 2021.

\bibitem{Xie_2018_Early}
L.~Xie, D.~Tao, and H.~Wei, ``Early expression detection via online multi-instance learning with nonlinear extension,'' \emph{IEEE Trans. Neural Networks Learn. Sys.}, vol.~30, pp. 1486--1496, 2018.

\bibitem{Zhang_2019_Multi}
Z.~Zhang, Y.~Nie, H.~Sun, Q.~Zhang, Q.~Lai, G.~Li, and M.~Xiao, ``Multi-view video synopsis via simultaneous object-shifting and view-switching optimization,'' \emph{IEEE Trans. Image Process.}, vol.~29, pp. 971--985, 2020.

\bibitem{Zhang_2024_novel}
H.~Zhang, B.~Li, S.-F. Su, W.~Yang, and L.~Xie, ``A novel hybrid transformer-based framework for solar irradiance forecasting under incomplete data scenarios,'' \emph{IEEE Trans. Ind. Inf.}, 2024.

\bibitem{Shah_2023_Multi}
K.~Shah, A.~Shah, C.~P. Lau, C.~M. de~Melo, and R.~Chellappa, ``Multi-view action recognition using contrastive learning,'' in \emph{Proc. Winter Conf. Appl. Comput. Vis.}, 2023, pp. 3381--3391.

\bibitem{Das_2023_ViewCLR}
S.~Das and M.~S. Ryoo, ``Viewclr: Learning self-supervised video representation for unseen viewpoints,'' in \emph{Proc. Winter Conf. Appl. Comput. Vis.}, 2023, pp. 5573--5583.

\bibitem{Xue_2023_Learning}
Z.~S. Xue and K.~Grauman, ``Learning fine-grained view-invariant representations from unpaired ego-exo videos via temporal alignment,'' \emph{Proc. Adv. Neural Inf. Process. Syst.}, vol.~36, pp. 53\,688--53\,710, 2023.

\bibitem{Gao_2021_View}
L.~Gao, Y.~Ji, K.~Gedamu, X.~Zhu, X.~Xu, and H.~T. Shen, ``View-invariant human action recognition via view transformation network (vtn),'' \emph{IEEE Trans. Multimedia}, vol.~24, pp. 4493--4503, 2021.

\bibitem{Xu_2022_Probabilistic}
J.~Xu, G.~Chen, N.~Zhou, W.-S. Zheng, and J.~Lu, ``Probabilistic temporal modeling for unintentional action localization,'' \emph{IEEE Trans. Image Process.}, vol.~31, pp. 3081--3094, 2022.

\bibitem{Xu_2023_Conditional}
X.~Xu, Y.~Wang, L.~Wang, B.~Yu, and J.~Jia, ``Conditional temporal variational autoencoder for action video prediction,'' \emph{Int. J. Comput. Vision}, vol. 131, no.~10, pp. 2699--2722, 2023.

\bibitem{Tong_2023_Probabilistic}
J.~Tong, L.~Xie, and K.~Zhang, ``Probabilistic decomposition transformer for time series forecasting,'' in \emph{Proc. SIAM Int. Conf. Data Min.}, 2023, pp. 478--486.

\bibitem{DeGeest_2016_Online}
R.~De~Geest, E.~Gavves, A.~Ghodrati, Z.~Li, C.~Snoek, and T.~Tuytelaars, ``Online action detection,'' in \emph{Proc. Eur. Conf. Comput. Vis.}, 2016, pp. 269--284.

\bibitem{Xu_2019_Temporal}
M.~Xu, M.~Gao, Y.-T. Chen, L.~S. Davis, and D.~J. Crandall, ``Temporal recurrent networks for online action detection,'' in \emph{Proc. IEEE Int. Conf. Comput. Vis.}, 2019, pp. 5532--5541.

\bibitem{Eun_2020_Learning}
H.~Eun, J.~Moon, J.~Park, C.~Jung, and C.~Kim, ``Learning to discriminate information for online action detection,'' in \emph{Proc. IEEE Conf. Comput. Vis. Pattern Recognit.}, 2020, pp. 809--818.

\bibitem{Kim_2021_Temporally}
Y.~H. Kim, S.~Nam, and S.~J. Kim, ``Temporally smooth online action detection using cycle-consistent future anticipation,'' \emph{Pattern Recogn.}, vol. 116, p. 107954, 2021.

\bibitem{An_2023_MiniROAD}
J.~An, H.~Kang, S.~H. Han, M.-H. Yang, and S.~J. Kim, ``Miniroad: Minimal rnn framework for online action detection,'' in \emph{Proc. IEEE Int. Conf. Comput. Vis.}, 2023, pp. 10\,341--10\,350.

\bibitem{Wang_2021_OadTR}
X.~Wang, S.~Zhang, Z.~Qing, Y.~Shao, Z.~Zuo, C.~Gao, and N.~Sang, ``Oadtr: Online action detection with transformers,'' in \emph{Proc. IEEE Int. Conf. Comput. Vis.}, 2021, pp. 7565--7575.

\bibitem{Xu_2021_Long}
M.~Xu, Y.~Xiong, H.~Chen, X.~Li, W.~Xia, Z.~Tu, and S.~Soatto, ``Long short-term transformer for online action detection,'' \emph{Adv. Neural Inf. Process. Syst.}, vol.~34, pp. 1086--1099, 2021.

\bibitem{Chen_2022_GateHUB}
J.~Chen, G.~Mittal, Y.~Yu, Y.~Kong, and M.~Chen, ``Gatehub: Gated history unit with background suppression for online action detection,'' in \emph{Proc. IEEE Conf. Comput. Vis. Pattern Recognit.}, 2022, pp. 19\,925--19\,934.

\bibitem{Zhao_2022_Real}
Y.~Zhao and P.~Krähenbühl, ``Real-time online video detection with temporal smoothing transformers,'' in \emph{Proc. Eur. Conf. Comput. Vis.}, 2022, pp. 485--502.

\bibitem{Wang_2023_Memory}
J.~Wang, G.~Chen, Y.~Huang, L.~Wang, and T.~Lu, ``Memory-and-anticipation transformer for online action understanding,'' in \emph{Proc. IEEE Int. Conf. Comput. Vis.}, 2023, pp. 13\,824--13\,835.

\bibitem{Liu_2024_HCM}
S.~Liu, J.~Cheng, Z.~Xia, Z.~Xi, Q.~Hou, and Z.~Dong, ``Hcm: Online action detection with hard video clip mining,'' \emph{IEEE Trans. Multimedia}, vol.~26, pp. 3626--3639, 2024.

\bibitem{Siddiqui_2024_DVANet}
N.~Siddiqui, P.~Tirupattur, and M.~Shah, ``Dvanet: Disentangling view and action features for multi-view action recognition,'' in \emph{Proc. AAAI Conf. Artif. Intell.}, vol.~38, no.~5, 2024, p. 4873–4881.

\bibitem{Marsella_2021_Adversarial}
A.~Marsella, G.~Goyal, and F.~Odone, ``Adversarial feature refinement for cross-view action recognition,'' in \emph{Proc. ACM Symp. Appl. Computing}, 2021, pp. 1046--1054.

\bibitem{Wang_2018_Dividing}
D.~Wang, W.~Ouyang, W.~Li, and D.~Xu, ``Dividing and aggregating network for multi-view action recognition,'' in \emph{Proc. Eur. Conf. Comput. Vis.}, 2018, pp. 451--467.

\bibitem{Haresh_2021_Learning}
S.~Haresh, S.~Kumar, H.~Coskun, S.~N. Syed, A.~Konin, Z.~Zia, and Q.-H. Tran, ``Learning by aligning videos in time,'' in \emph{Proc. IEEE Conf. Comput. Vis. Pattern Recognit.}, 2021, pp. 5548--5558.

\bibitem{Piergiovanni_2021_Recognizing}
A.~Piergiovanni and M.~S. Ryoo, ``Recognizing actions in videos from unseen viewpoints,'' in \emph{Proc. IEEE Conf. Comput. Vis. Pattern Recognit.}, 2021, pp. 4124--4132.

\bibitem{Ghoddoosian_2022_Weakly}
R.~Ghoddoosian, I.~Dwivedi, N.~Agarwal, C.~Choi, and B.~Dariush, ``Weakly-supervised online action segmentation in multi-view instructional videos,'' in \emph{Proc. IEEE Conf. Comput. Vis. Pattern Recognit.}, 2022, pp. 13\,780--13\,790.

\bibitem{Zhou_2019_Recognizing}
J.~Zhou and T.~Komuro, ``Recognizing fall actions from videos using reconstruction error of variational autoencoder,'' in \emph{Proc. IEEE Int. Conf. Image Process.}, 2019, pp. 3372--3376.

\bibitem{Srivastava_2021_variational}
A.~Srivastava, O.~Dutta, J.~Gupta, S.~Agarwal, and P.~AP, ``A variational information bottleneck based method to compress sequential networks for human action recognition,'' in \emph{Proc. Winter Conf. Appl. Comput. Vis.}, 2021, pp. 2745--2754.

\bibitem{Guo_2022_Action2video}
C.~Guo, X.~Zuo, S.~Wang, X.~Liu, S.~Zou, M.~Gong, and L.~Cheng, ``Action2video: Generating videos of human 3d actions,'' \emph{Int. J. Comput. Vision}, vol. 130, no.~2, pp. 285--315, 2022.

\bibitem{Panousis_2021_Variational}
K.~P. Panousis, S.~Chatzis, and S.~Theodoridis, ``Variational conditional dependence hidden markov models for skeleton-based action recognition,'' in \emph{Proc. Lect. Notes Comput. Sci.}, 2021, pp. 67--80.

\bibitem{Wei_2023_Unsupervised}
P.~Wei, L.~Kong, X.~Qu, Y.~Ren, Z.~Xu, J.~Jiang, and X.~Yin, ``Unsupervised video domain adaptation for action recognition: A disentanglement perspective,'' \emph{Proc. Adv. Neural Inf. Process. Syst.}, vol.~36, pp. 17\,623--17\,642, 2023.

\bibitem{Kingma_2013_Auto}
D.~P. Kingma and M.~Welling, ``Auto-encoding variational bayes,'' \emph{arXiv preprint arXiv:1312.6114}, 2013.

\bibitem{Doersch_2016_Tutorial}
C.~Doersch, ``Tutorial on variational autoencoders,'' \emph{arXiv preprint arXiv:1606.05908}, 2016.

\bibitem{BenShabat_2021_IKEA}
Y.~Ben-Shabat, X.~Yu, F.~Saleh, D.~Campbell, C.~Rodriguez-Opazo, H.~Li, and S.~Gould, ``The ikea asm dataset: Understanding people assembling furniture through actions, objects and pose,'' in \emph{Proc. Winter Conf. Appl. Comput. Vis.}, 2021, pp. 847--859.

\bibitem{Vaquette_2017_DAily}
G.~Vaquette, A.~Orcesi, L.~Lucat, and C.~Achard, ``The daily home life activity dataset: a high semantic activity dataset for online recognition,'' in \emph{Proc. IEEE Int. Conf. Auto. Face Gesture Recognit.}, 2017, pp. 497--504.

\bibitem{Kuehne_2014_Language}
H.~Kuehne, A.~Arslan, and T.~Serre, ``The language of actions: Recovering the syntax and semantics of goal-directed human activities,'' in \emph{Proc. IEEE Conf. Comput. Vis. Pattern Recognit.}, 2014, pp. 780--787.

\bibitem{Wang_2016_Temporal}
L.~Wang, Y.~Xiong, Z.~Wang, Y.~Qiao, D.~Lin, X.~Tang, and L.~Van~Gool, ``Temporal segment networks: Towards good practices for deep action recognition,'' in \emph{Proc. Eur. Conf. Comput. Vis.}, 2016, pp. 20--36.

\bibitem{Carreira_2017_Quo}
J.~Carreira and A.~Zisserman, ``Quo vadis, action recognition? a new model and the kinetics dataset,'' in \emph{Proc. IEEE Conf. Comput. Vis. Pattern Recognit.}, 2017, pp. 6299--6308.

\bibitem{Contributors_2020_OpenMMLabs}
M.~Contributors, ``Openmmlab's next generation video understanding toolbox and benchmark,'' \url{https://github.com/open-mmlab/mmaction2}, 2020.

\bibitem{Kingma_2014_Adam}
D.~P. Kingma and J.~Ba, ``Adam: A method for stochastic optimization,'' \emph{arXiv preprint arXiv:1412.6980}, 2014.

\bibitem{Loshchilov_2016_Sgdr}
I.~Loshchilov and F.~Hutter, ``Sgdr: Stochastic gradient descent with warm restarts,'' \emph{arXiv preprint arXiv:1608.03983}, 2016.

\bibitem{Zhang_2013_Cross}
Z.~Zhang, C.~Wang, B.~Xiao, W.~Zhou, S.~Liu, and C.~Shi, ``Cross-view action recognition via a continuous virtual path,'' in \emph{Proc. IEEE Conf. Comput. Vis. Pattern Recognit.}, 2013, pp. 2690--2697.

\bibitem{Ullah_2021_Conflux}
A.~Ullah, K.~Muhammad, T.~Hussain, and S.~W. Baik, ``Conflux lstms network: A novel approach for multi-view action recognition,'' \emph{Neurocomputing}, vol. 435, pp. 321--329, 2021.

\bibitem{Das_2020_VPN}
S.~Das, S.~Sharma, R.~Dai, F.~Bremond, and M.~Thonnat, ``Vpn: Learning video-pose embedding for activities of daily living,'' in \emph{Proc. Eur. Conf. Comput. Vis.}, 2020, pp. 72--90.

\end{thebibliography}
\bibliographystyle{IEEEtran}

\end{document}